\documentclass[runningheads]{llncs}

\usepackage[mobile]{eccv}

\usepackage{eccvabbrv}
\usepackage{xcolor}
\usepackage{amssymb}
\usepackage{graphicx}
\usepackage{booktabs}
\usepackage{colortbl}
\usepackage{multirow}
\definecolor{mygray}{gray}{.9}
\usepackage[accsupp]{axessibility}  % Improves PDF readability for those with disabilities.

\definecolor{eccvblue}{rgb}{0.12,0.49,0.85}

\usepackage[pagebackref,breaklinks,colorlinks,citecolor=eccvblue]{hyperref}

\usepackage{orcidlink}

\usepackage{makecell}

\newcommand{\best}[1]{{{\textcolor{red}{#1}}}}
\newcommand{\second}[1]{{\textcolor{blue}{{#1}}}}

\usepackage{xspace}
\newcommand{\NAME}{MambaIR\xspace}

\usepackage[capitalize]{cleveref}
\crefname{section}{Sec.}{Secs.}
\Crefname{section}{Section}{Sections}
\Crefname{table}{Table}{Tables}
\crefname{table}{Tab.}{Tabs.}

\begin{document}

\title{\raisebox{-0.19em}{\includegraphics[height=1.1em]{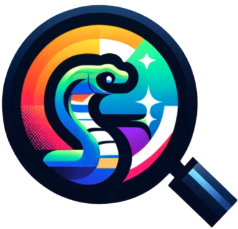}}MambaIR: A Simple Baseline for Image Restoration with State-Space Model} 

% \title{Sigma\raisebox{-0.19em}{\includegraphics[height=1.1em]{figs/snake.png}}: Siamese Mamba Network for Multi-Modal Semantic Segmentation} 

\titlerunning{MambaIR}

\author{Hang Guo\inst{1, 4, \star} 
\and 
Jinmin Li\inst{1, \star}
\and Tao Dai\inst{2, \dagger}
\and \\ 
Zhihao Ouyang\inst{3, 4} 
\and Xudong Ren\inst{1} 
\and Shu-Tao Xia\inst{1,5}}

\authorrunning{Guo et al.}

\institute{$^1$Tsinghua Shenzhen International Graduate School, Tsinghua University \\
$^2$College of Computer Science and Software Engineering, Shenzhen University \\
$^3$ByteDance Inc. \qquad $^4$Aitist.ai \qquad $^5$Peng Cheng Laboratory \\ 
\email{\{cshguo, daitao.edu\}@gmail.com, \{ljm22,rxd21\}@mails.tsinghua.edu.cn\\
zhihao.ouyang@bytedance.com, xiast@sz.tsinghua.edu.cn
}
}

\setcounter{footnote}{0}
\renewcommand{\thefootnote}{}
\footnotetext[0]{$^\star$Equal contribution.}
\footnotetext[0]{$^\dagger$Corresponding author: Tao Dai (\tt{daitao.edu@gmail.com})}

\maketitle

\begin{abstract}

Recent years have seen significant advancements in image restoration, largely attributed to the development of modern deep neural networks, such as CNNs and Transformers. However, existing restoration backbones often face the dilemma between global receptive fields and efficient computation, hindering their application in practice. Recently, the Selective Structured State Space Model, especially the improved version Mamba, has shown great potential for long-range dependency modeling with linear complexity, which offers a way to resolve the above dilemma. However, the standard Mamba still faces certain challenges in low-level vision such as local pixel forgetting and channel redundancy. In this work, we introduce a simple but effective baseline, named MambaIR, which introduces both local enhancement and channel attention to improve the vanilla Mamba. In this way, our MambaIR takes advantage of the local pixel similarity and reduces the channel redundancy. Extensive experiments demonstrate the superiority of our method, for example, MambaIR outperforms SwinIR by up to 0.45dB on image SR, using similar computational cost but with a global receptive field. Code is available at \url{https://github.com/csguoh/MambaIR}.

 
\end{abstract}

\section{Introduction}
\label{sec:intro}

Image restoration, aiming to reconstruct a high-quality image from a given low-quality input, is a long-standing problem in computer vision and further has a wide range of sub-problems such as super-resolution, image denoising, \etc. With the introduction of modern deep learning models such as CNNs~\cite{DnCNN,dong2014learning,lim2017enhanced,zhang2018residual,dai2019second} and Transformers~\cite{chen2021pre,liang2021swinir,chen2023activating,li2023grl,chen2023dual}, state-of-the-art performance has continued to be refreshed in the past few years.

To some extent, the increasing performance of deep restoration models largely stems from the increasing network receptive field. First, a large receptive field allows the network to capture information from a wider region, enabling it to refer to more pixels to facilitate the reconstruction of the anchor pixel. Second, with a larger receptive field, the restoration network can extract higher-level patterns and structures in the image, which can be crucial for some structure preservation tasks such as image denoising. Finally, Transformer-based restoration methods which possess larger receptive fields experimentally outperform CNN-based methods, and the recent work~\cite{chen2023activating} also points out that activating more pixels usually leads to better restoration results.

\begin{figure*}[!t]
\centering
\includegraphics[width=\textwidth]{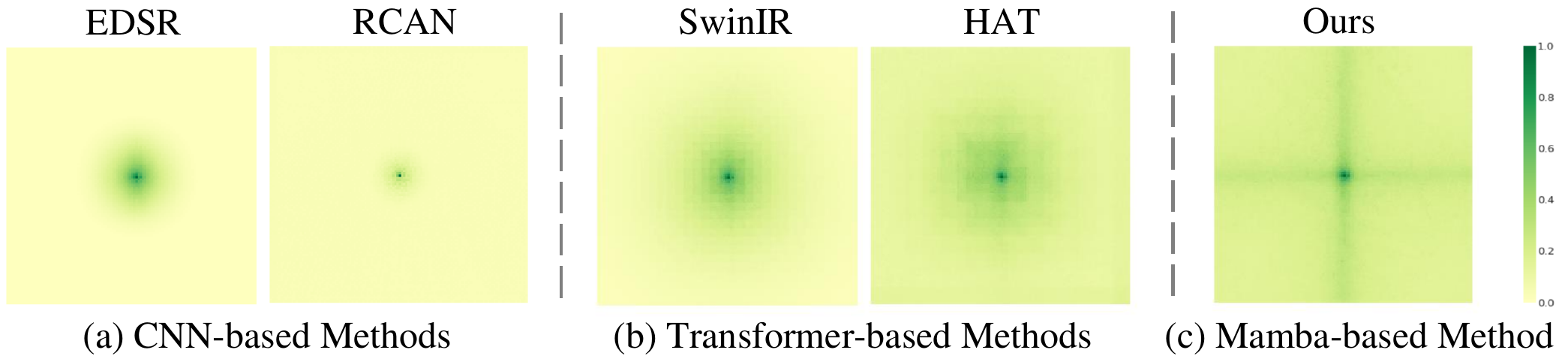}
\caption{The Effective Receptive Field (ERF) visualization~\cite{luo2016understanding,ding2022scaling} for EDSR~\cite{lim2017enhanced}, RCAN~\cite{zhang2018image}, SwinIR~\cite{liang2021swinir}, HAT~\cite{chen2023activating}, and the proposed \NAME. A larger ERF is indicated by a more extensively distributed dark area. The proposed \NAME achieves a significant global effective receptive field.}
\label{fig:erf}
\end{figure*}

Despite possessing many attractive properties, it appears that there exists an inherent \textit{choice dilemma} between global receptive fields and efficient computation for current image restoration backbones. For CNN-based restoration networks~\cite{lim2017enhanced,zhang2018residual}, although the effective receptive field is limited (as shown in~\cref{fig:erf}(a)), it is appropriate for resource-constrained device deployments due to the favorable efficiency of convolution parallel operations. By contrast, the Transformer-based image restoration methods usually set the number of tokens to the image resolution~\cite{liang2021swinir,chen2021pre,chen2023activating}, therefore, despite the global receptive field, directly using the standard Transformer~\cite{vaswani2017attention} will come at an unacceptable quadratic computational complexity. Moreover, employing some efficient attention techniques such as shifted window attention~\cite{liu2021swin} for image restoration, usually comes at the expense of a globally effective receptive field (as shown in ~\cref{fig:erf}(b)), and does not intrinsically escape out of the trade-off between a global receptive field and efficient computation.

Recently, structured state-space sequence models (S4), especially the improved version Mamba, have emerged as an efficient and effective backbone for constructing deep networks~\cite{gu2021efficiently,smith2022simplified,fu2022hungry,mehta2022long,gu2023mamba}. This development hints at a potential solution to balancing global receptive field and computational efficiency in image restoration. In detail, the discretized state space equations in Mamba can be formalized into a recursive form and can model very long-range dependencies when equipped with specially designed structured reparameterization~\cite{gu2020hippo}. This means that Mamba-based restoration networks can naturally activate more pixels, thus improving the reconstruction quality. Furthermore, the parallel scan algorithm~\cite{gu2023mamba} renders Mamba to process each token in a parallel fashion, facilitating efficient training on modern hardware such as GPU. The above promising properties motivate us to explore the potential of Mamba to achieve efficient long-range modeling for image restoration networks.

However, the standard Mamba~\cite{gu2023mamba}, which is designed for 1D sequential data in NLP, is not a natural fit for image restoration scenarios. First, since Mamba processes flattened 1D image sequences in a recursive manner, it can result in spatially close pixels being found at very distant locations in the flattened sequences, resulting in the problem of local pixel forgetting. Second, due to the requirement to memorize the long sequence dependencies, the number of hidden states in the state space equations is typically large, which can lead to channel redundancy, thus hindering the learning of critical channel representations.

To address the above challenges, we introduce \NAME, a simple but very effective benchmark model, to adapt Mamba for image restoration. \NAME is formulated with three principal stages. Specifically, the \textbf{1)Shallow Feature Extraction} stage employs a simple convolution layer to extract the shallow feature. Then the \textbf{2)Deep Feature Extraction} stage performs with several stacked Residual State Space Blocks (RSSBs). As the core component of our \NAME, the RSSB is designed with local convolution to mitigate local pixel forgetting when applying the vanilla Mamba to 2D images, and it is also equipped with channel attention to reduce channel redundancy caused by the excessive hidden state number. We also employ the learnable factor to control the skip connection within each RSSB. Finally, the \textbf{3)High-Quality Image Reconstruction} stage aggregates both shallow and deep features to produce a high-quality output image. Through possessing both a global effective receptive field as well as linear computational complexity, our \NAME serves as a new alternative for image restoration backbones.

In short, our main contributions can be summarized as follows: 
\begin{itemize}
    \item We are the first work to adapt state space models for low-level image restoration via extensive experiments to formulate \NAME, which acts as a simple but effective alternative for CNN- and Transformer-based methods.
    \item We propose the Residue State Space Block (RSSB) which can boost the power of the standard Mamba with local enhancement and channel redundancy reduction.
    \item Extensive experiments on various tasks demonstrate our \NAME outperforms other strong baselines to provide a powerful and promising backbone solution for image restoration. 
\end{itemize}

\section{Related Work}

\subsection{Image Restoration}

Image restoration has been significantly advanced since the introduction of deep learning by several pioneering works, such as SRCNN~\cite{dong2014learning} for image super-resolution, DnCNN~\cite{DnCNN} for image denoising,  ARCNN~\cite{dong2015compression} for JPEG compression artifact reduction, \etc. Early attempts usually elaborate CNNs with techniques such as residual connection~\cite{kim2016accurate,cavigelli2017cas}, dense connection~\cite{wang2018esrgan,zhang2018residual} and others~\cite{lai2017deep,wei2021unsupervised,fu2019jpeg,dai2019second} to improve model representation ability. Despite the success, CNN-based restoration methods typically face challenges in effectively modeling global dependencies. As transformer have proven its effectiveness in multiple tasks, such as time series~\cite{liu2024taming}, 3D cloud~\cite{zha2023instance,zha2024towards}, and multi-modal~\cite{gao2024inducing,bai2023badclip,gao2024energy,zhang2024vision}, using transformer for image restoration appears promising. Despite the global receptive field, transformer still faces specific challenges from the quadratic computational complexity of the self-attention~\cite{vaswani2017attention}. To address this, IPT~\cite{chen2021pre} divides one image into several small patches and processes each patch independently with self-attention. SwinIR~\cite{liang2021swinir} further introduces shifted window attention~\cite{liu2021swin} to improve the performance. In addition, progress continues to be made in designing efficient attention for restoration~\cite{zhang2023accurate,chen2023activating,li2021efficient,chen2023dual,zamir2022restormer,chen2023recursive, sun2023safmn,guo2023adaptir, zhang2024parameter, chen2022simple}. Nonetheless, efficient attention design usually comes at the expense of global receptive fields, and the dilemma of the trade-off between efficient computation and global modeling is not essentially resolved.

\subsection{State Space Models}

State Space Models (SSMs)~\cite{gu2021combining,gu2021efficiently,smith2022simplified}, stemming from classics control theory~\cite{kalman1960new}, are recently introduced to deep learning as a competitive backbone for state space transforming. The promising property of linearly scaling with sequence length in long-range dependency modeling has attracted great interest from searchers.  For example, the Structured State-Space Sequence model (S4)~\cite{gu2021efficiently} is a pioneer work for the deep state-space model in modeling the long-range dependency. Later, S5 layer~\cite{smith2022simplified} is proposed based on S4 and introduces MIMO SSM and efficient parallel scan. Moreover, H3~\cite{fu2022hungry} achieves promising results that nearly fill the performance gap between SSMs and Transformers in natural language. ~\cite{mehta2022long} further improve S4 with gating units to obtain the Gated State Space layer to boost the capability. More recently, Mamba~\cite{gu2023mamba}, a data-dependent SSM with selective mechanism and efficient hardware design, outperforms Transformers on 
natural language and enjoys linear scaling with input length. Moreover, there are also pioneering works that adopt Mamba to vision tasks such as image classification~\cite{liu2024vmamba,zhu2024vision}, video understanding~\cite{wang2023selective,li2024videomamba}, biomedical image segmentation~\cite{ma2024u,xing2024segmamba} and others~\cite{islam2023efficient,nguyen2022s4nd, hu2024zigma,zha2024lcm,qin2024mambavc}. In this work, we explore the potential of Mamba to image restoration with restoration-specific designs to serve as a simple but effective baseline for future work.

\section{Methodology}

% In this section, we begin with a description of the preliminaries of the SSM. Then we give an overview of our simple but effective benchmark model \NAME, followed by a detailed description of the proposed Residule State-Space Block (RSSB). Finally, we provide a comprehensive discussion of the difference between our \NAME and previous methods.

\subsection{Preliminaries}

The recent advancements of the class of structured state-space sequence models (S4) are largely inspired by the continuous linear time-invariant (LTI) systems, which maps a 1-dimensional function or sequence $x(t) \in \mathbb{R} \rightarrow y(t) \in \mathbb{R}$ through an implicit latent state $h(t)\in \mathbb{R}^{N}$. Formally, this system can be formulated as a linear ordinary differential equation (ODE):

\begin{equation}
\begin{aligned}
\label{eq:ssm}
    h'(t)&={\rm \textbf{A}}h(t)+{\rm \textbf{B}}x(t),\\
    y(t)&={\rm \textbf{C}}h(t)+{\rm \textbf{D}}x(t),
\end{aligned}
\end{equation}

\noindent
where $N$ is the state size, ${\rm \textbf{A}} \in \mathbb{R}^{N\times N}$, ${\rm \textbf{B}} \in \mathbb{R}^{N \times 1}$, ${\rm \textbf{C}} \in \mathbb{R}^{1\times N}$, and ${\rm \textbf{D}} \in \mathbb{R}$.

After that, the discretization process is typically adopted to integrate \cref{eq:ssm} into practical deep learning algorithms. Specifically, denote $\rm \Delta$ as the timescale parameter to transform the continuous parameters ${{\rm \textbf{A}}}$, ${{\rm \textbf{B}}}$ to discrete parameters $\overline{{\rm \textbf{A}}}$, $\overline{{\rm \textbf{B}}}$. The commonly used method for discretization is the zero-order hold (ZOH) rule, which is defined as follows:

\begin{equation}
\begin{aligned}
    \overline{\rm \textbf{A}} &= {\rm exp}({\rm {\Delta \textbf{A}}}),\\
    \overline{\rm \textbf{B}}&=({\rm {\Delta \textbf{A}}})^{-1}({\rm exp(\textbf{A})}-\textbf{I})\cdot {\rm \Delta \textbf{B}}.
\end{aligned}
\end{equation}

After the discretization, the discretized version of \cref{eq:ssm} with step size $\rm \Delta$ can be rewritten in the following RNN form:

\begin{equation}
\begin{aligned}
\label{eq:discret-ssm}
    h_k&=\overline{\rm \textbf{A}}h_{k-1}+\overline{\rm \textbf{B}}x_k,\\
    y_k&={\rm \textbf{C}}h_k+{\rm \textbf{D}}x_k.
\end{aligned}
\end{equation}

Furthermore, the~\cref{eq:discret-ssm} can also be mathematically equivalently transformed into the following CNN form:

\begin{equation}
\begin{aligned}
\label{eq:cnn-form}
\overline{\rm \textbf{K}}&\triangleq(\mathrm{\textbf{C}} \overline{\mathrm{\textbf{B}}},{\rm{\textbf{C}}}\overline{\rm \textbf{A}}\overline{\rm \textbf{B}},\cdots,{\rm{\textbf{C}}}{\overline{\rm \textbf{A}}}^{L-1}\overline{\rm \textbf{B}}),\\
{\rm \textbf{y}}&={\rm \textbf{x}} \circledast \overline{\rm \textbf{K}},
\end{aligned}
\end{equation}

\noindent
where $L$ is the length of the input sequence, $\circledast$ denotes convolution operation, and $\overline{\rm \textbf{K}} \in \mathbb{R}^L$ is a structured convolution kernel.

The recent advanced state-space model, Mamba~\cite{gu2023mamba}, have further improved 
$\overline{\rm \textbf{B}}$, ${\rm \textbf{C}}$ and $\rm \Delta$ to be input-dependent, thus allowing for a dynamic feature representation. The intuition of Mamba for image restoration lies in its development on the advantages of  S4 model. Specifically, Mamba shares the same recursive form of ~\cref{eq:discret-ssm}, which enables the model to memorize ultra-long sequences so that more pixels can be activated to aid restoration. At the same time,  the parallel scan algorithm~\cite{gu2023mamba} allows Mamba to enjoy the same advantages of parallel processing as \cref{eq:cnn-form}, thus facilitating efficient training.

%\subsection{MambaIR: Taming Mamba for Super-resolution}

\subsection{Overall Architecture}

\begin{figure*}[!t]
\centering
\includegraphics[width=\textwidth]{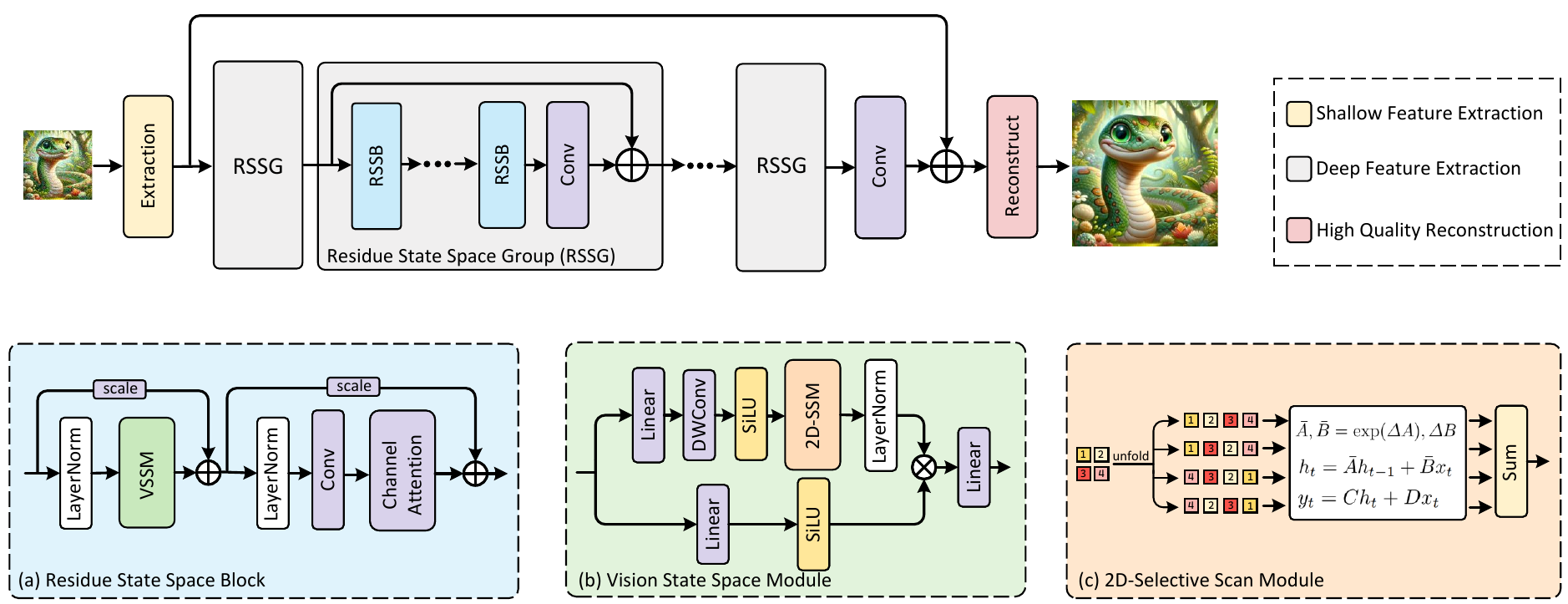}
\caption{The overall network architecture of our \NAME, as well as the (a) Residual State-Space Block (RSSB), the (b) Vision State-Space Module (VSSM), and the (c) 2D Selective Scan Module (2D-SSM).}
\label{fig:pipeline}
\end{figure*}

As shown in \cref{fig:pipeline},  our \NAME consists of three stages: shallow feature extraction, deep feature extraction, and high-quality reconstruction. Given a low-quality (LQ) input image $I_{LQ} \in \mathbb{R}^{H \times W \times 3}$, we first employ a $3\times 3$ convolution layer from the shallow feature extraction to generate the shallow feature $F_S \in \mathbb{R}^{H \times W \times C}$, where $H$ and $W$ represent the height and width of the input image, and $C$ is the number of channels. Subsequently, the shallow feature $F_S$ undergoes the deep feature extraction stage to acquire the deep feature $F_D^l \in \mathbb{R}^{H \times W \times C}$ at the $l$-th layer, $l \in \{1,2,\cdots L\}$. This stage is stacked by multiple Residual State-Space Groups (RSSGs), with each RSSG containing several Residue State-Space Blocks (RSSBs). Moreover, an additional convolution layer is introduced at the end of each group to refine features extracted from RSSB. Finally, we use the element-wise sum to obtain the input of the high-quality reconstruction stage $F_R = F^L_D + F_S$, which is used to reconstruct the high-quality (HQ) output image $I_{HQ}$.

\subsection{Residual State-Space Block}

The block design in previous Transformer-based restoration networks~\cite{liang2021swinir,zhang2023accurate,chen2023activating,chen2023dual} mainly follow the \texttt{Norm} $\rightarrow$ \texttt{Attention} $\rightarrow$ \texttt{Norm} $\rightarrow$ \texttt{MLP} flow. Although Attention and SSM can both model global dependencies, however, we find these two modules behave differently (see \textit{supplementary material} for more details) and simply replacing Attention with SSM can only obtain sub-optimal results. Therefore, it is promising to tailor a brand-new block structure for Mamba-based restoration networks.

To this end, we propose the Residual State-Space Block (RSSB) to adapt the SSM block for restoration. As shown in \cref{fig:pipeline}(a), given the input deep feature $F_D^l \in \mathbb{R}^{H\times W \times C}$, we first use the LayerNorm (LN) followed by the Vision State-Space Module (VSSM)~\cite{liu2024vmamba} to capture the spatial long-term dependency. Moreover, we also use learnable scale factor $s \in \mathbb{R}^C$  to control the information from skip connection:

\begin{equation}
Z^l = \mathrm{VSSM}(\mathrm{LN}(F_D^l))+s\cdot F_D^l.
\end{equation}

\begin{figure*}[!t]
\centering
\includegraphics[width=0.9\textwidth]{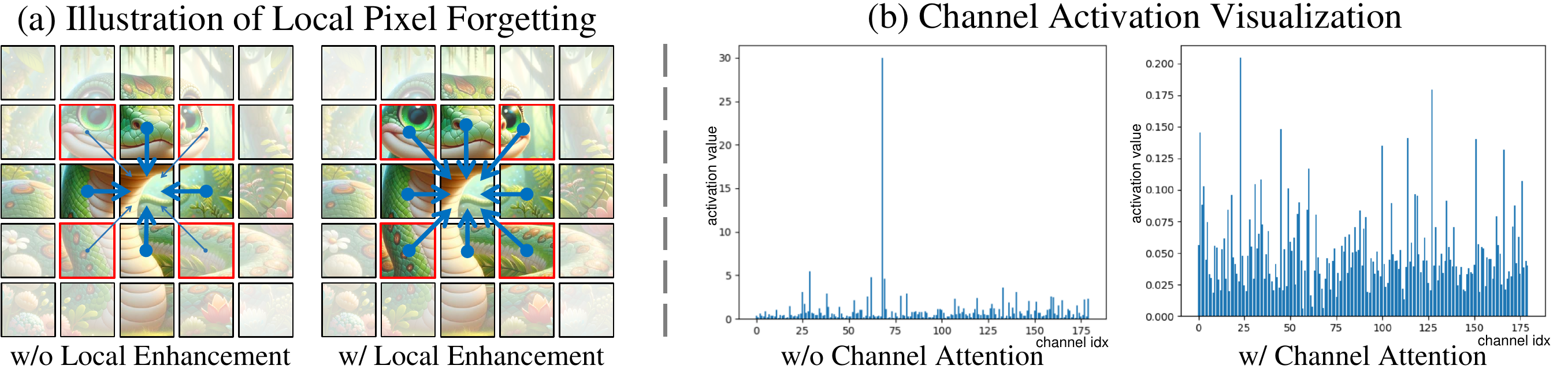}
\caption{(a) Without using local enhancement will cause spatially close pixels (area in the red box) get forgotten in the flattened 1D sequence due to the long distance. (b) We use RELU and global average pooling on the VSSM outputs from the last layer to get the channel activation values. Most channels are not activated (\textit{i.e.}, channel redundancy) when channel attention is not used.}
\label{fig:channel-redundency}
\end{figure*}

Furthermore, since SSMs process flattened feature maps as 1D token sequences, the number of neighborhood pixels in the sequence is greatly influenced by the flattening strategy. For example, when employing the four-direction unfolding strategy of ~\cite{liu2024vmamba}, only four nearest neighbors are available to the anchor pixel (see ~\cref{fig:channel-redundency}(a)), \textit{i.e.}, some spatially close pixels in 2D feature map are instead distant from each other in the 1D token sequence, and this over-distance can lead to local pixel forgetting. To this end, we introduce an additional local convolution after VSSM to help restore the neighborhood similarity. Specifically, we employ LayerNorm to first normalize the $Z^l$ and then use convolution layers to compensate for local features. In order to maintain efficiency, the convolution layer adopts the bottleneck structure, \textit{i.e.}, the channel is first compressed by a factor $\gamma$ to obtain features with the shape $\mathbb{R}^{H \times W \times \frac{C}{\gamma}}$, then we perform channel expansion to recover the original shape.

In addition, SSMs typically introduce a larger number of hidden states to memorize very long-range dependencies, and we visualize the activation results for different channels in ~\cref{fig:channel-redundency}(b) and find notable channel redundancy. To facilitate the expressive power of different channels, we introduce the Channel Attention (CA)~\cite{hu2018squeeze} to RSSB. In this way, SSMs can focus on learning diverse channel representations after which the critical channels are selected by subsequent channel attention, thus avoiding channel redundancy. At last, another tunable scale factor $s'\in \mathbb{R}^C$ is used in residual connection to acquire the final output $F_D^{l+1}$ of the RSSB. The above process can be formulated as:

\begin{equation}
    F_D^{l+1}= \mathrm{CA(Conv(LN}(Z^l))) + s' \cdot Z^l.
\end{equation}

\subsection{Vision State-Space Module}

To maintain efficiency, the Transformer-based restoration networks usually divide input into small patches~\cite{chen2021pre} or adopt shifted window attention~\cite{liang2021swinir}, hindering the interaction at the whole-image level. Motivated by the success of Mamba in long-range modeling with linear complexity, we introduce the Vision State-Space Module to image restoration.

The Vision State-Space Module (VSSM) can capture long-range dependencies with the state space equation, and the architecture of VSSM is shown in ~\cref{fig:pipeline}(b). Following~\cite{liu2024vmamba}, the input feature $X \in \mathbb{R}^{H \times W \times C}$ will go through two parallel branches. In the first branch, the feature channel is expanded to $\lambda C$ by a linear layer, where $\lambda$ is a pre-defined channel expansion factor, followed by a depth-wise convolution, SiLU~\cite{shazeer2020glu} activation function, together with the 2D-SSM layer and LayerNorm. In the second branch, the features channel is also expanded to $\lambda C$ with a linear layer followed by the SiLU activation function. After that, features from the two branches are aggregated with the Hadamard product. Finally, the channel number is projected back to $C$ to generate output $X_{out}$ with the same shape as input:

\begin{equation}
\begin{aligned}
&X_1 = \mathrm{LN(2D\text{-}SSM(SiLU(DWConv(Linear}(X))))),\\
&X_2 = \mathrm{SiLU(Linear}(X)),\\
&X_{out} = \mathrm{Linear}(X_1 \odot X_2),
\end{aligned}
\end{equation}

\noindent
where DWConv represents depth-wise convolution, and $\odot$ denotes the Hadamard product.

\subsection{2D Selective Scan Module}

The standard Mamba~\cite{gu2023mamba} causally processes the input data, and thus can only capture information within the scanned part of the data. This property is well suited for NLP tasks that involve a sequential nature but poses significant challenges when transferring to non-causal data such as images. To better utilize the 2D spatial information, we follow~\cite{liu2024vmamba} and introduce the 2D Selective Scan Module (2D-SSM). As shown in \cref{fig:pipeline}(c), the 2D image feature is flattened into a 1D sequence with scanning along four different directions: top-left to bottom-right, bottom-right to top-left, top-right to bottom-left, and bottom-left to top-right. Then the long-range dependency of each sequence is captured according to the discrete state-space equation. Finally, all sequences are merged using summation followed by the reshape operation to recover the 2D structure.

\subsection{Loss Function}
To make a fair comparison with previous works~\cite{zhang2018residual,liang2021swinir,zhang2023accurate}, we optimize our \NAME with $L_1$ loss for image SR, which can be formulated as:
\begin{equation}
\mathcal{L}=||I_{HQ}-I_{LQ}||_1,
\end{equation}

\noindent 
where $||\cdot||_1$ denotes the $L_1$ norm. For image denoising, we utilize the Charbonnier loss~\cite{charbonnier1994two} with $\epsilon = 10^{-3}$:
\begin{equation}
    \mathcal{L}=\sqrt{||I_{HQ}-I_{LQ}||^2+\epsilon^2}.
\end{equation}

\section{Experiences}

\subsection{Experimental Settings}

\noindent \textbf{Dataset and Evaluation.} 
Following the setup in previous works~\cite{liang2021swinir,zhang2023accurate}, we conduct experiments on various image restoration tasks, including image super-resolution (\textit{i.e.}, classic SR, lightweight SR, real SR) and image denoising (\textit{i.e.}, Gaussian color image denoising and real-world denoising), and JPEG compression artifact reduction (JPEG CAR). We employ DIV2K~\cite{timofte2017ntire} and Flickr2K~\cite{lim2017enhanced} to train classic SR models and use DIV2K only to train lightweight SR models. Moreover, we use Set5~\cite{bevilacqua2012low}, Set14~\cite{zeyde2012single}, B100~\cite{martin2001database}, Urban100~\cite{huang2015single}, and Manga109~\cite{matsui2017sketch} to evaluate the effectiveness of different SR methods. For gaussian color image denoising, we utilize DIV2K~\cite{timofte2017ntire}, Flickr2K~\cite{lim2017enhanced}, BSD500~\cite{arbelaez2010contour}, and WED~\cite{ma2016waterloo} as our training datasets. Our testing datasets for guassian color image denoising includes BSD68~\cite{martin2001database}, Kodak24~\cite{kao}, McMaster~\cite{zhang2011color}, and Urban100~\cite{huang2015single}. For real image denoising, 
we train our model with 320 high-resolution images from SIDD~\cite{abdelhamed2018high} datasets, and use the SIDD test set and DND ~\cite{plotz2017benchmarking} dataset for testing. 
Following~\cite{liang2021swinir,zhang2018residual}, we denote the model as MambaIR+ when self-ensemble strategy~\cite{lim2017enhanced} is used in testing. The performance is evaluated using PSNR and SSIM on the Y channel from the YCbCr color space. Due to page limit, the results of JPEG CAR are shown in the \textit{supplementary material}.

\noindent \textbf{Training Details.} 
In accordance with previous works~\cite{liang2021swinir,chen2023activating,zhang2023accurate}, we perform data augmentation by applying horizontal flips and random rotations of $90^\circ, 180^\circ$, and $270^\circ$. Additionally, we crop the original images into $64 \times 64$ patches for image SR and $128 \times 128$ patches for image denoising during training. For image SR, we use the pre-trained weights from the $\times$2 model to initialize those of $\times$3 and $\times$4 and halve the learning rate and total training iterations to reduce training time~\cite{lim2017enhanced}. To ensure a fair comparison, we adjust the training batch size to 32 for image SR and 16 for image denoising. We employ the Adam~\cite{kingma2014adam} as the optimizer for training our MambaIR with $\beta_1 = 0.9, \beta_2 = 0.999$. The initial learning rate is set at $2 \times 10^{-4}$ and is halved when the training iteration reaches specific milestones. Our MambaIR model is trained with 8 NVIDIA V100 GPUs.

\subsection{Ablation Study}

% For ablation experiments, we train our models for image super-resolution ($\times$2) based on DIV2K datasets and evaluate on Urban100.

\noindent \textbf{Effects of different designs of RSSB.} 
As the core component, the RSSB can improve Mamba with restoration-specific priors. In this section, we ablate different components of the RSSB. The results, presented in \cref{tab:ablation-rssb}, indicate that (1) applying 1D scan on flattened images can lead to local pixel forgetting, and the utilization of simple convolution layers can effectively enhance the local interaction. (2) Without using additional convolution and channel attention, \textit{i.e.}, directly employing off-the-shelf Mamba for restoration, can only obtain sub-optimal results, which also supports our previous analysis. (3) Replacing Conv+ChanelAttention with MLP, whose resulted structure will be similar to Transformer, also leads to unfavorable results, indicating that although both SSMs and Attention have the global modeling ability, the behavior of these two modules is different and thus accustomed block structure should be considered for further improvements.

\begin{figure}[!tb]  
\begin{minipage}[t]{0.48\linewidth}
        \centering
        \captionsetup{width=0.96\linewidth}
        \captionof{table}{Ablation experiments for different design choices of RSSB.}
       \label{tab:ablation-rssb}
        \setlength{\tabcolsep}{1.8pt}
        \scalebox{0.799}{
       \begin{tabular}{l|cccccc}
\toprule
settings        & Set5  & Set14  & Urban100 \\ \midrule
(1)remove Conv     & 38.48 & 34.54  & 34.04  \\
(2)remove Conv+CA  &38.55 & 34.64 & 34.06  \\
(3)replace with MLP  &38.55 & 34.68 & 34.22 \\ \bottomrule
\end{tabular}%
        }
    \end{minipage}
    \begin{minipage}[t]{0.48\linewidth}
    \centering
    \captionsetup{width=0.9\linewidth}
        \captionof{table}{Ablation experiments for different scan modes in VSSM.}
        \label{tab:ablation-direction}
        \setlength{\tabcolsep}{1.8pt}
        \scalebox{0.799}{
\begin{tabular}{@{}l|cccccc@{}}
\toprule
scan mode     &  Set5  & Set14  & Urban100 \\ \midrule
one-direction  &38.53 & 34.63  &  34.06       \\
two-direction &38.56 & 34.60 &  33.96       \\
baseline     &38.57 & 34.67  & 34.15        \\ \bottomrule
\end{tabular}%
}
    \end{minipage}
\end{figure}

% \begin{table}[!t]
% \centering
% \caption{Ablation experiments for different design choices of RSSB.}
% \label{tab:ablation-rssb}
% \setlength{\tabcolsep}{2pt}
% \scalebox{0.92}{
% \begin{tabular}{l|cc|ccccc}
% \toprule
% settings        & \#param & MACs & Set5  & Set14 & B100 & Urban100 & Manga109 \\ \midrule
% (0)baseline     &  20.6M & 706G & 38.57 & 34.67 & 32.58 & 34.15 & 40.28 \\
% (1)+w/o Conv    & 13.6M & 527G  & 38.48 & 34.54 & 32.56 & 34.04 & 40.20 \\
% (2)+w/o Conv+CA  & 13.5M & 526G &38.55 & 34.64 & 32.57 & 34.06 & 40.14 \\
% (3)+replace with MLP  & 18.2M & 646G &38.55 & 34.68 & 32.59 & 34.22 & 40.13 \\ \bottomrule
% \end{tabular}
% }
% \end{table}

% \begin{table}[!t]
% \centering
% \caption{Ablation experiments for different scan modes in VSSM.}
% \label{tab:ablation-direction}
% \setlength{\tabcolsep}{2pt}
% \scalebox{0.92}{
% \begin{tabular}{@{}l|cc|ccccc@{}}
% \toprule
% scan mode     & \#param & MACs& Set5  & Set14 & B100  & Urban100 & Manga109 \\ \midrule
% one-direction & 17.7M & 506G &38.53 & 34.63 & 32.58 &  34.06   & 40.31     \\
% two-direction & 18.6M & 572G &38.56 & 34.60 & 32.56 &  33.96   & 40.14     \\
% baseline      & 20.6M & 706G &38.57 & 34.67 & 32.58 & 34.15    & 40.28     \\ \bottomrule
% \end{tabular}%
% }
% \end{table}

\begin{figure*}[!t]
\centering
\includegraphics[width=0.95\textwidth]{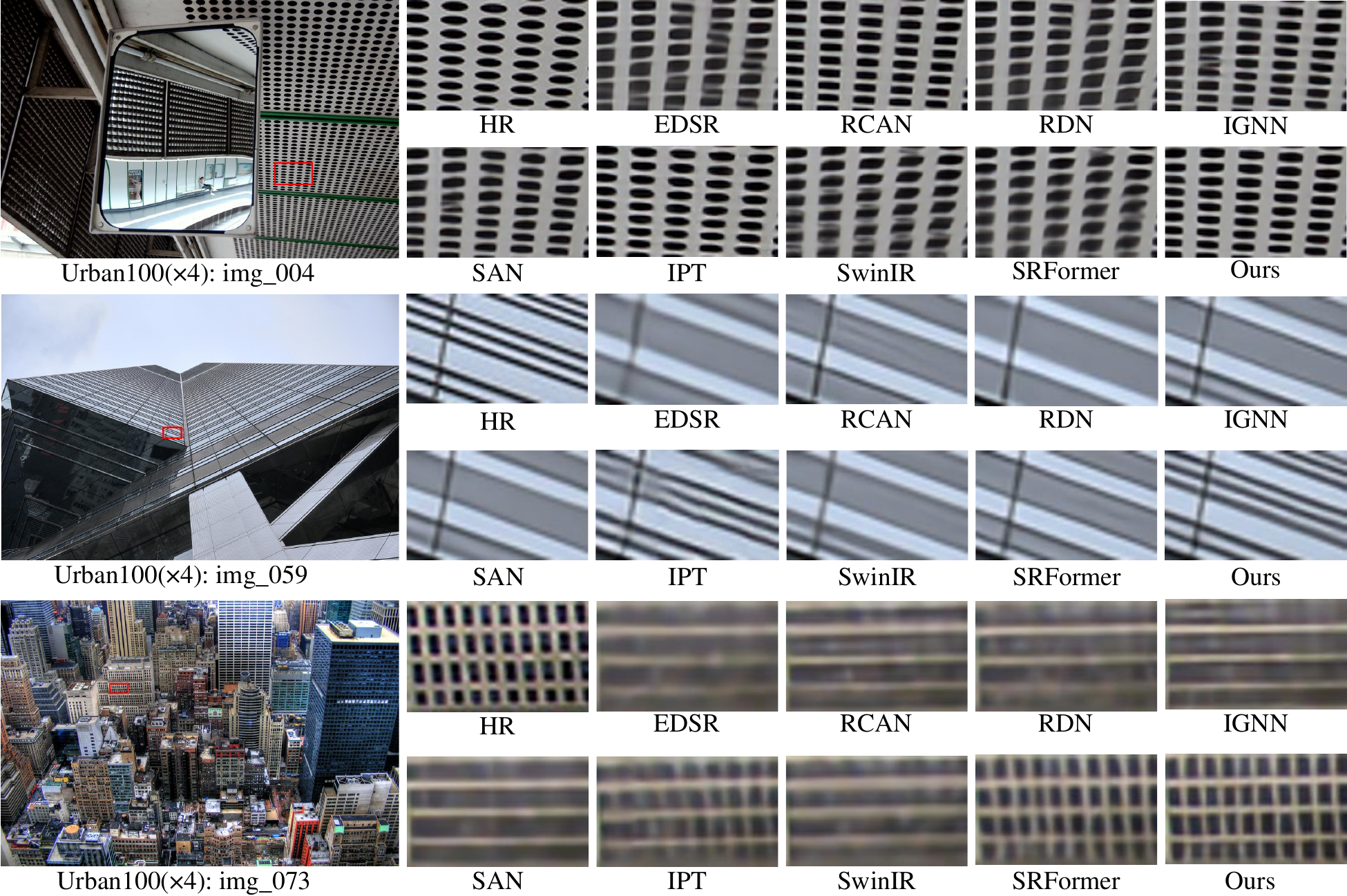}
\caption{Qualitative comparison of our \NAME with CNN and Transformer based methods on \underline{\textbf{classic image SR}} with scale $\times$4.}
\label{fig:visual-sr}
\end{figure*}

\begin{table*}[!t]
\centering
\caption{Quantitative comparison on \underline{\textbf{classic image super-resolution}} with state-of-the-art methods. The best and the second best results are in \best{red} and \second{blue}.}
\label{tab:classicSR}
\setlength{\tabcolsep}{2pt}
\scalebox{0.92}{
\begin{tabular}{@{}l|c|cc|cc|cc|cc|cc@{}}
\toprule
 & & \multicolumn{2}{c|}{\textbf{Set5}} &
  \multicolumn{2}{c|}{\textbf{Set14}} &
  \multicolumn{2}{c|}{\textbf{BSDS100}} &
  \multicolumn{2}{c|}{\textbf{Urban100}} &
  \multicolumn{2}{c}{\textbf{Manga109}} \\
\multirow{-2}{*}{Method} & \multirow{-2}{*}{scale} & PSNR  & SSIM   & PSNR  & SSIM   & PSNR  & SSIM   & PSNR  & SSIM   & PSNR  & SSIM   \\ \midrule
EDSR~\cite{lim2017enhanced}   & $\times 2$ & 38.11 & 0.9602 & 33.92 & 0.9195 & 32.32 & 0.9013 & 32.93 & 0.9351 & 39.10 & 0.9773 \\
RCAN~\cite{zhang2018image}   & $\times 2$ & 38.27 & 0.9614 & 34.12 & 0.9216 & 32.41 & 0.9027 & 33.34 & 0.9384 & 39.44 & 0.9786 \\
SAN~\cite{dai2019second}    & $\times 2$ & 38.31 & 0.9620 & 34.07 & 0.9213 & 32.42 & 0.9028 & 33.10 & 0.9370 & 39.32 & 0.9792 \\
HAN~\cite{niu2020single}    & $\times 2$ & 38.27 & 0.9614 & 34.16 & 0.9217 & 32.41 & 0.9027 & 33.35 & 0.9385 & 39.46 & 0.9785 \\
IGNN~\cite{zhou2020cross} & $\times$2 &38.24 & 0.9613 & 34.07 & 0.9217 & 32.41 & 0.9025 & 33.23 & 0.9383 & 39.35 & 0.9786\\
CSNLN~\cite{mei2020image}  & $\times 2$ & 38.28 & 0.9616 & 34.12 & 0.9223 & 32.40 & 0.9024 & 33.25 & 0.9386 & 39.37 & 0.9785 \\
NLSA~\cite{mei2021image}   & $\times 2$ & 38.34 & 0.9618 & 34.08 & 0.9231 & 32.43 & 0.9027 & 33.42 & 0.9394 & 39.59 & 0.9789 \\
ELAN~\cite{zhang2022efficient}   & $\times 2$ & 38.36 & 0.9620 & 34.20 & 0.9228 & 32.45 & 0.9030 & 33.44 & 0.9391 & 39.62 & 0.9793 \\
IPT~\cite{chen2021pre} & $\times 2$ & 38.37 & - &34.43 &-& 32.48&-& 33.76& -& -& - \\
SwinIR~\cite{liang2021swinir} & $\times 2$ & 38.42 & 0.9623 & 34.46 & 0.9250 & 32.53 & 0.9041 & 33.81 & 0.9427 & 39.92 & 0.9797 \\
SRFormer~\cite{zhou2023srformer} & $\times 2$ &  38.51 &0.9627&34.44&0.9253&32.57 &0.9046 &34.09& \best{0.9449} &40.07 &0.9802\\
\NAME & $\times 2$ &\second{38.57}&\second{0.9627}&\second{34.67}&\best{0.9261}&\second{32.58}&\second{0.9048}&\second{34.15}&\second{0.9446}&\second{40.28}&\second{0.9806} \\
MambaIR+   & $\times 2$ &\best{38.60}&\best{0.9628}&\best{34.69}&\second{0.9260}&\best{32.60}&\best{0.9048}&\best{34.17}&0.9443&\best{40.33}&\best{0.9806} \\     \midrule
EDSR~\cite{lim2017enhanced} & $\times$3 & 
34.65 & 0.9280 & 30.52 & 0.8462 & 29.25 & 0.8093 & 28.80 & 0.8653 & 34.17 & 0.9476\\
RCAN~\cite{zhang2018image} & $\times$3 &
34.74 & 0.9299 & 30.65 & 0.8482 & 29.32 & 0.8111 & 29.09 & 0.8702 & 34.44 & 0.9499\\
SAN~\cite{dai2019second} & $\times$3 &
34.75 & 0.9300 & 30.59 & 0.8476 & 29.33 & 0.8112 & 28.93 & 0.8671 & 34.30 & 0.9494\\
HAN~\cite{niu2020single} & $\times$3 &
34.75 & 0.9299 & 30.67 & 0.8483 & 29.32 & 0.8110 & 29.10 & 0.8705 & 34.48 & 0.9500\\
IGNN~\cite{zhou2020cross} & $\times$3 &
34.72 & 0.9298 & 30.66 & 0.8484 & 29.31 & 0.8105 & 29.03 & 0.8696 & 34.39 & 0.9496\\
CSNLN~\cite{mei2020image} & $\times$3 &
34.74 & 0.9300 & 30.66 & 0.8482 & 29.33 & 0.8105 & 29.13 & 0.8712 & 34.45 & 0.9502\\
NLSA~\cite{mei2021image} & $\times$3 &
34.85 & 0.9306 & 30.70 & 0.8485 & 29.34 & 0.8117 & 29.25 & 0.8726 & 34.57 & 0.9508\\
ELAN~\cite{zhang2022efficient} & $\times$3 &
34.90 & 0.9313 & 30.80 & 0.8504 & 29.38 & 0.8124 & 29.32 & 0.8745 & 34.73 & 0.9517\\
IPT~\cite{chen2021pre} & $\times$3 &
34.81 & - & 30.85 & - & 29.38 & - & 29.49 & - & - & - \\
SwinIR~\cite{liang2021swinir} & $\times$3 &
34.97 & 0.9318 & 30.93 & 0.8534 & 29.46 & 0.8145 & 29.75 & 0.8826 & 35.12 & 0.9537\\
SRformer~\cite{zhou2023srformer} & $\times$3 &
35.02& 0.9323 & 30.94 & \second{0.8540} & 29.48 & 0.8156 & \best{30.04} & 0.8865 & 35.26 & 0.9543\\
\NAME & $\times$3 & \second{35.08} & \second{0.9323} & \second{30.99} & {0.8536} & \second{29.51} & \second{0.8157} & {29.93} & \second{0.8841} & \second{35.43} & \second{0.9546} \\
MambaIR+ & $\times$3 &  \best{35.13} & \best{0.9326} & \best{31.06} & \best{0.8541} & \best{29.53} & \best{0.8162} & \second{29.98} & \best{0.8838} & \best{35.55} & \best{0.9549}
\\ \midrule
EDSR~\cite{lim2017enhanced}   & $\times 4$ & 32.46 & 0.8968 & 28.80 & 0.7876 & 27.71 & 0.7420 & 26.64 & 0.8033 & 31.02 & 0.9148 \\
RCAN~\cite{zhang2018image}     & $\times 4$ & 32.63 & 0.9002 & 28.87 & 0.7889 & 27.77 & 0.7436 & 26.82 & 0.8087 & 31.22 & 0.9173 \\
SAN~\cite{dai2019second}      & $\times 4$ & 32.64 & 0.9003 & 28.92 & 0.7888 & 27.78 & 0.7436 & 26.79 & 0.8068 & 31.18 & 0.9169 \\
HAN~\cite{niu2020single}      & $\times 4$ & 32.64 & 0.9002 & 28.90 & 0.7890 & 27.80 & 0.7442 & 26.85 & 0.8094 & 31.42 & 0.9177 \\
IGNN~\cite{zhou2020cross}     & $\times$4 &32.57 & 0.8998 & 28.85 & 0.7891 & 27.77 & 0.7434 & 26.84 & 0.8090 & 31.28 & 0.9182\\
CSNLN~\cite{mei2020image}     & $\times 4$ & 32.68 & 0.9004 & 28.95 & 0.7888 & 27.80 & 0.7439 & 27.22 & 0.8168 & 31.43 & 0.9201 \\
NLSA~\cite{mei2021image}      & $\times 4$ & 32.59 & 0.9000 & 28.87 & 0.7891 & 27.78 & 0.7444 & 26.96 & 0.8109 & 31.27 & 0.9184 \\
ELAN~\cite{zhang2022efficient}   & $\times 4$ & 32.75 & 0.9022 & 28.96 & 0.7914 & 27.83 & 0.7459 & 27.13 & 0.8167 & 31.68 & 0.9226 \\
IPT ~\cite{chen2021pre}          & $\times 4$ & 32.64 &-& 29.01 &- &27.82 &-& 27.26& -& -& - \\
SwinIR~\cite{liang2021swinir}    & $\times 4$ & 32.92 & 0.9044 & 29.09 & 0.7950 & 27.92 & 0.7489 & 27.45 & 0.8254 & 32.03 & 0.9260 \\
SRFormer~\cite{zhou2023srformer} & $\times 4$ & 32.93 & 0.9041 & 29.08 &0.7953 &27.94 & 0.7502 & 27.68 & \best{0.8311} & 32.21 & 0.9271 \\
\NAME   & $\times 4$ &\second{33.03}&\second{0.9046}&\second{29.20}&\second{0.7961}&\second{27.98}&\second{0.7503}&\second{27.68}&0.8287&\second{32.32}&\second{0.9272} \\
MambaIR+   & $\times 4$ &\best{33.13}&\best{0.9054}&\best{29.25}&\best{0.7971}&\best{28.01}&\best{0.7510}&\best{27.80}&\second{0.8303}&\best{32.48}&\best{0.9281} \\  \bottomrule
\end{tabular}%
}
\end{table*}

\noindent \textbf{Effects of Different Scan Modes in VSSM.}
To allow Mamba to process 2D images, the feature map needs to be flattened before being iterated by the state-space equation. Therefore, the unfolding strategy is particularly important. In this work, we follow~\cite{liu2024vmamba} which uses scans in four different directions to generate scanned sequences. Here, we ablate different scan modes to study the effects, the results are shown in \cref{tab:ablation-direction}. Compared with one-direction (top-left to bottom-right) and two-direction (top-left to bottom-right, bottom-right to top-left), using four directions of scanning allows the anchor pixel to perceive a wider range of neighborhoods, thus achieving better results. We also include other ablation experiments, such as the layer number of RSSBs, please see \textit{supplementary material} for more analysis.

\subsection{Comparison on Image Super-Resolution}

\noindent \textbf{Classic Image Super-Resolution.}
% 定量，定性，效率对比
\cref{tab:classicSR} shows the quantitative results between \NAME and state-of-the-art super-resolution methods. Thanks to the significant global receptive field, our proposed \NAME achieves the best performance on almost all five benchmark datasets for all scale factors. For example, our Mamba-based baseline outperforms the Transformer-based benchmark model SwinIR by 0.41dB on Manga109 for $\times 2$ scale, demonstrating the prospect of Mamba for image restoration. We also give visual comparisons in~\cref{fig:visual-sr}, and  our method can facilitate the reconstruction of sharp edges and natural textures. 
% We also give a comparison of the model complexity, see the \textit{supplementary material} for more details.
% Moreover, the LAM visualization~\cite{gu2021interpreting} shown in~\cref{fig:lam} demonstrates our \NAME can utilize more pixels during restoration due to the powerful long-range modeling capability of Mamba.

\begin{figure}[!t]
\centering
\includegraphics[width=0.45\textwidth]{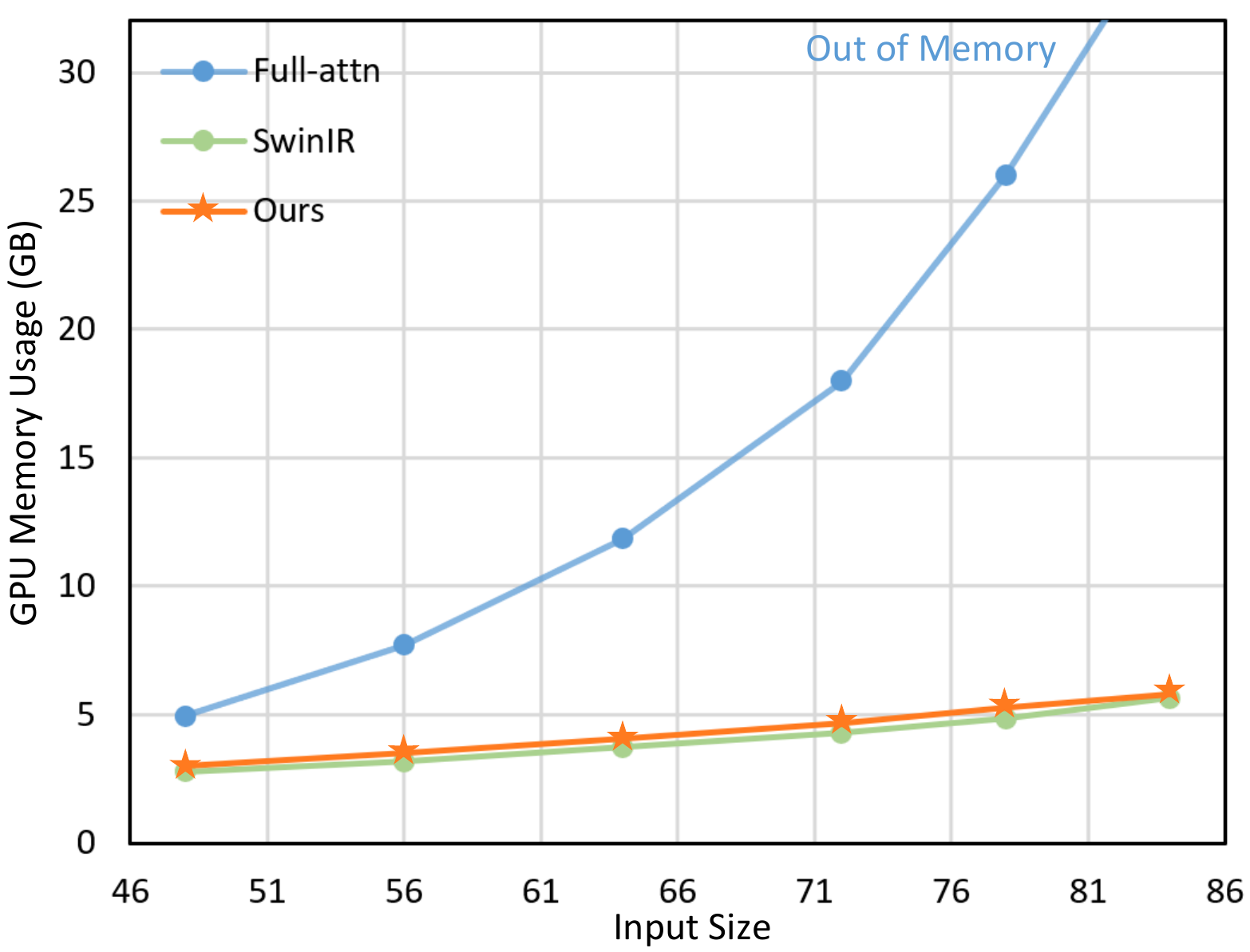}
\caption{Computational complexity comparison with different input scales. We set the standard attention~\cite{vaswani2017attention} which has a global receptive field as baseline, and denote it as "Full-attn". We adjust the model to ensure the GPU usage is roughly similar at the beginning, and then scale the input resolution from $48 \times 48$ to $84 \times 84$.}
\label{fig:gpu-memo}
\end{figure}

\noindent \textbf{Model Complexity Comparison.}
We give comparison results of computational complexity with varying input sizes in ~\cref{fig:gpu-memo}. As one can see, our method is far more efficient than the full-attention baseline~\cite{vaswani2017attention} and exhibits linear complexity with input resolution which is similar to the efficient attention techniques such as SwinIR. These observations above suggest that out \NAME has similar scale properties as shifted window attention, while possessing a global receptive field similar to standard full attention.

\noindent \textbf{Lightweight Image Super-Resolution.} 
To demonstrate the scalability of our method, we train the MambaIR-light model and compare it with state-of-the-art lightweight image SR methods. Following previous works~\cite{zhou2023srformer,luo2020latticenet}, we also report the number of parameters (\#param) and MACs (upscaling a low-resolution image to $1280 \times 720$ resolution). \cref{tab:lightSR} shows the results. It can be seen that our MambaIR-light outperforms SwinIR-light~\cite{liang2021swinir} by up to 0.34dB PSNR on the $\times 4$ scale Manga109 dataset with similar parameters and MACs. The performance results demonstrate the scalability and efficiency of our method.

\begin{table*}[!t]
\centering
\caption{Quantitative comparison on \underline{\textbf{lightweight image super-resolution}} with state-of-the-art methods.}
\label{tab:lightSR}
\setlength{\tabcolsep}{2pt}
\scalebox{0.75}{
\begin{tabular}{@{}l|c|c|c|cc|cc|cc|cc|cc@{}}
\toprule
 & & & & \multicolumn{2}{c|}{\textbf{Set5}} &
  \multicolumn{2}{c|}{\textbf{Set14}} &
  \multicolumn{2}{c|}{\textbf{BSDS100}} &
  \multicolumn{2}{c|}{\textbf{Urban100}} &
  \multicolumn{2}{c}{\textbf{Manga109}} \\
\multirow{-2}{*}{Method} & \multirow{-2}{*}{scale}& \multirow{-2}{*}{\#param}& \multirow{-2}{*}{MACs} & PSNR  & SSIM   & PSNR  & SSIM   & PSNR  & SSIM   & PSNR  & SSIM   & PSNR  & SSIM   \\ \midrule
CARN~\cite{ahn2018fast} & $\times$2  & 1,592K & 222.8G
& 37.76
& 0.9590
& 33.52
& 0.9166
& 32.09
& 0.8978
& 31.92
& 0.9256
& 38.36
& 0.9765
\\
IMDN~\cite{hui2019lightweight}& $\times$2  & 694K & 158.8G
& 38.00
& 0.9605
& 33.63
& 0.9177
& 32.19
& 0.8996
& 32.17
& 0.9283
& {38.88}
& {0.9774}
\\
LAPAR-A~\cite{li2020lapar} & $\times$2  & 548K & 171.0G
& 38.01
& 0.9605
& 33.62
& 0.9183
& 32.19
& 0.8999
& 32.10
& 0.9283
& 38.67
& 0.9772
\\
LatticeNet~\cite{luo2020latticenet} & $\times$2  & 756K & 169.5G
& {38.13}
& 0.9610
& {33.78}
& {0.9193}
& {32.25}
& {0.9005}
& {32.43}
& {0.9302}
& -
& -
\\
SwinIR-light~\cite{liang2021swinir}& $\times$2  & 910K & {122.2G} %
& \best{38.14}
& \best{0.9611}
& \second{33.86}
& \second{0.9206}
& \second{32.31}
& \second{0.9012}
& \second{32.76}
& \second{0.9340}
& \second{39.12}
& \best{0.9783}
\\
Ours & $\times$2  & 905K & 167.1G %
& \second{38.13}
& \second{0.9610}
& \best{33.95}
& \best{0.9208}
& \best{32.31}
& \best{0.9013}
& \best{32.85}
& \best{0.9349}
& \best{39.20}
& \second{0.9782}
\\
\midrule
CARN~\cite{ahn2018fast} & $\times$3  & 1,592K  & 118.8G
& 34.29
& 0.9255
& 30.29
& 0.8407
& 29.06
& 0.8034
& 28.06
& 0.8493
& 33.50 
& 0.9440
\\ 
IMDN~\cite{hui2019lightweight} & $\times$3  & 703K  & 71.5G 
& 34.36
& 0.9270
& 30.32
& 0.8417
& 29.09
& 0.8046
& 28.17
& 0.8519
& {33.61}
& {0.9445}
\\ 
LAPAR-A~\cite{li2020lapar} & $\times$3  & 544K & 114.0G
& 34.36
& 0.9267
& 30.34
& 0.8421
& 29.11
& 0.8054
& 28.15
& 0.8523
& 33.51
& 0.9441
\\
LatticeNet~\cite{luo2020latticenet} & $\times$3  & 765K & 76.3G 
& {34.53}
& {0.9281}
& {30.39}
& {0.8424}
& {29.15}
& {0.8059}
& {28.33}
& {0.8538}
& -
& -
\\
SwinIR-light~\cite{liang2021swinir} & $\times$3  & 918K & {55.4G} %
& \second{34.62}
& \best{0.9289}
& \second{30.54}
& \best{0.8463}
& \second{29.20}
& \second{0.8082}
& \second{28.66}
& \second{0.8624}
& \second{33.98}
& \second{0.9478}
\\ 
Ours & $\times$3  & 913K & {74.5G} %
& \best{34.63}
& \second{0.9288}
& \best{30.54}
& \second{0.8459}
& \best{29.23}
& \best{0.8084}
& \best{28.70}
& \best{0.8631}
& \best{34.12}
& \best{0.9479}
\\
\midrule
CARN~\cite{ahn2018fast} & $\times$4  & 1,592K & 90.9G
& 32.13
& 0.8937
& 28.60
& 0.7806
& 27.58
& 0.7349
& 26.07 
& 0.7837
& {30.47}
& {0.9084}
\\
IMDN~\cite{hui2019lightweight}& $\times$4  & 715K & 40.9G
& 32.21
& 0.8948
& 28.58
& 0.7811
& 27.56
& 0.7353 
& 26.04
& 0.7838
& 30.45
& 0.9075
\\
LAPAR-A~\cite{li2020lapar} & $\times$4  & 659K & 94.0G
& 32.15
& 0.8944
& 28.61
& 0.7818
& 27.61
& 0.7366
& 26.14
& 0.7871
& 30.42
& 0.9074
\\
LatticeNet~\cite{luo2020latticenet} & $\times$4  & 777K & 43.6G
& {32.30}
& {0.8962}
& {28.68}
& {0.7830}
& {27.62}
& {0.7367}
& {26.25}
& {0.7873}
& -
& -
\\
SwinIR-light~\cite{liang2021swinir} & $\times$4  & 930K & {31.8G} %
& \best{32.44}
& \second{0.8976}
& \best{28.77}
& \best{0.7858}
& \best{27.69}
& \best{0.7406}
& \second{26.47}
& \second{0.7980}
& \second{30.92}
& \best{0.9151}
\\
Ours  & $\times$4  &  924K & 42.3G %
& \second{32.42}
& \best{0.8977}
& \second{28.74}
& \second{0.7847}
& \second{27.68}
& \second{0.7400}
& \best{26.52}
& \best{0.7983}
& \best{30.94}
& \second{0.9135}
\\
\bottomrule
\end{tabular}%
}
\end{table*}

\begin{figure*}[!t]
\centering
\includegraphics[width=\textwidth]{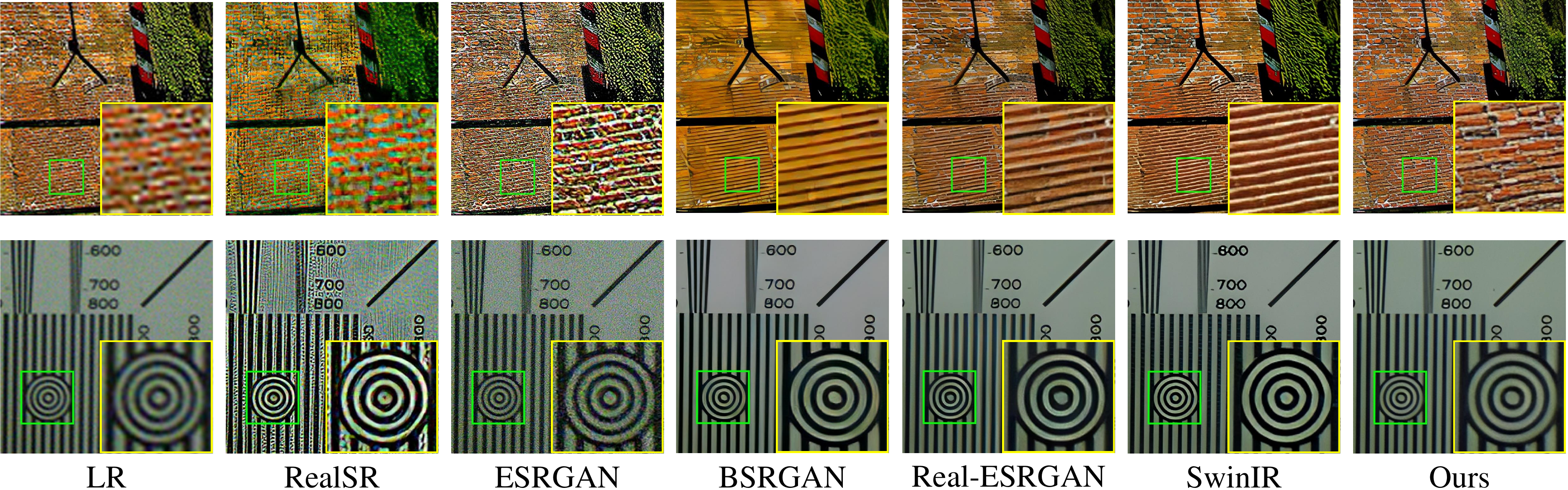}
\caption{Qualitative comparison with RealSR~\cite{ji2020real}, ESRGAN~\cite{wang2018esrgan}, BSRGAN~\cite{zhang2021designing}, Real-ESRGAN~\cite{wang2021real}, and SwinIR~\cite{liang2021swinir} on \underline{\textbf{real image super-resolution}} with scale $\times$4.}
\label{fig:realSR}
\end{figure*}

\noindent \textbf{Real-world Image Super-resolution.} 
We also investigate the performance of the network for real-world image restoration. We follow the training protocol in ~\cite{chen2023activating} to train our MambaIR-real model. Since there are no ground-truth images for this task, only the visual comparison is given in \cref{fig:realSR}. Compared with the other methods, our \NAME exhibits a notable advancement in resolving fine details and texture preservation, demonstrating the robustness of our method.

\subsection{Comparison on Image Denoising}
\begin{figure*}[!t]
\centering
\includegraphics[width=0.98\textwidth]{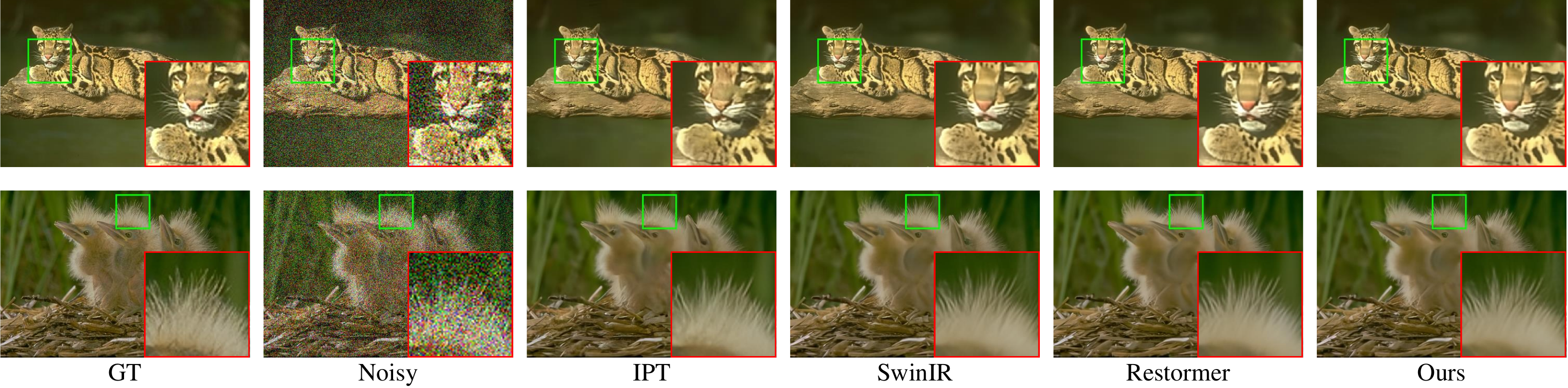}
\caption{Qualitative comparison of our \NAME with other methods on \underline{\textbf{color image denoising}} task with noise level level $\sigma$=50.}
\label{fig:visual-dn}
\end{figure*}

\noindent \textbf{Gaussian Color Image Denoising.}
The results of gaussian color image denoising are shown in ~\cref{tab:guassian-denoise}. Following~\cite{DnCNN,DRUNet}, the compared noise levels include 15, 25 and 50. As one can see, our model achieves the best performance on most datasets. In particular, it surpasses the SwinIR~\cite{liang2021swinir} by even 0.48dB with $\sigma$=50 on the Urban100 dataset.  We also give a visual comparison in \cref{fig:visual-dn}. Thanks to the global receptive field, our \NAME can achieve better structure preservation, leading to clearer edges and natural shapes.

\begin{table*}[!t]
\centering
\caption{Quantitative comparison on \underline{\textbf{gaussian color image denoising}} with state-of-the-art methods.}
\label{tab:guassian-denoise}
\setlength{\tabcolsep}{2pt}
\scalebox{0.9}{
\begin{tabular}{l|ccc|ccc|ccc|ccc}
\toprule
\multirow{2}{*}{Method} &  \multicolumn{3}{c|}{\textbf{BSD68}} & \multicolumn{3}{c|}{\textbf{Kodak24}} & \multicolumn{3}{c|}{\textbf{McMaster}} & \multicolumn{3}{c}{\textbf{Urban100}}\\
&$\sigma$=15 & $\sigma$=25 & $\sigma$=50 &$\sigma$=15 & $\sigma$=25 & $\sigma$=50 &$\sigma$=15 & $\sigma$=25 & $\sigma$=50 &$\sigma$=15 & $\sigma$=25 & $\sigma$=50\\
\midrule
IRCNN~\cite{IRCNN} &
33.86 & 31.16 & 27.86 & 34.69 & 32.18 & 28.93 & 34.58 & 32.18 & 28.91 & 33.78 & 31.20 & 27.70\\
FFDNet~\cite{FFDNet} &
33.87 & 31.21 & 27.96 & 34.63 & 32.13 & 28.98 & 34.66 & 32.35 & 29.18 & 33.83 & 31.40 & 28.05\\
DnCNN~\cite{DnCNN} &
33.90 & 31.24 & 27.95 & 34.60 & 32.14 & 28.95 & 33.45 & 31.52 & 28.62 & 32.98 & 30.81 & 27.59\\
DRUNet~\cite{DRUNet}&
34.30 & 31.69 & 28.51 & 35.31 & 32.89 & 29.86 & 35.40 & 33.14 & 30.08 & 34.81 & 32.60 & 29.61\\
SwinIR~\cite{liang2021swinir} &
\second{34.42} & 31.78 & 28.56 & 35.34 & 32.89 & 29.79 & 35.61 & 33.20 & 30.22 & 35.13 & 32.90 & 29.82\\
Restormer~\cite{zamir2022restormer} & 34.40 & \second{31.79} & \second{28.60} & \best{35.47} & \best{33.04} & \best{30.01} & \second{35.61} & \second{33.34} & \second{30.30} & \second{35.13} & \second{32.96} & \second{30.02} \\ 
MambaIR &
\best{34.48} & \best{32.24} & \best{28.66} & \second{35.42} & \second{32.99} & \second{29.92} & \best{35.70} &\best{33.43} & \best{30.35} & \best{35.37} & \best{33.21} & \best{30.30}\\
\bottomrule
\end{tabular} 
}
\end{table*}

\noindent \textbf{Real Image Denoising.}
We further turn to the real image denoising task to evaluate the robustness of \NAME when facing real-world degradation. Following~\cite{zamir2022restormer}, we adopt the progressive training strategy for fair comparison. The results, shown in \cref{tab:real-denoise}, suggest that our method achieves comparable performance with existing state-of-the-art models Restormer~\cite{wang2022uformer} and outperforms other methods such as Uformer~\cite{wang2022uformer} by 0.12dB PSNR on SIDD dataset, indicating the ability of our method in real image denoising.

\begin{table*}[!t]
\centering
\caption{Quantitative comparison on the \underline{\textbf{real image denosing}} task.}
\label{tab:real-denoise}
\setlength{\tabcolsep}{2pt}
\scalebox{0.9}{
\begin{tabular}{l|cc|cc|cc|cc|cc|cc}
\toprule
&  \multicolumn{2}{c|}{ DeamNet~\cite{ren2021adaptive}} & \multicolumn{2}{c|}{ MPRNet~\cite{zamir2021multi} } & \multicolumn{2}{c|}{ DAGL~\cite{mou2021dynamic} } & \multicolumn{2}{c|}{ Uformer~\cite{wang2022uformer} }& \multicolumn{2}{c|}{ Restormer~\cite{zamir2022restormer} } & \multicolumn{2}{c}{Ours} \\
\multirow{-2}{*}{Dataset} & PSNR & SSIM & PSNR & SSIM & PSNR & SSIM & PSNR & SSIM & PSNR & SSIM & PSNR & SSIM \\
\midrule
SIDD & 39.47 & 0.957 & 39.71 & 0.958 & 38.94 & 0.953 & {39.77} & {0.959}& \best{40.02} & \best{0.960} & \second{39.89} & \second{0.960}\\
DND &  39.63 & 0.953 & 39.80 & 0.954 & 39.77 & 0.956 & {39.96} & {0.956} & \second{40.03} & \best{0.956} & \best{40.04} & \second{0.956} \\ \bottomrule
\end{tabular}%
}
\end{table*}

\section{Conclusion}
In this work, we explore for the first time the power of the recent advanced state space model, \textit{i.e.}, Mamba, for image restoration, to help resolve the dilemma of trade-off between efficient computation and global effective receptive field. Specifically, we introduce the local enhancement to mitigate the neighborhood pixel forgetting problem from the flattening strategy and propose channel attention to reduce channel redundancy. Extensive experiments on multiple restoration tasks demonstrate our MambaIR serves as a simple but effective state-space model for image restoration.

\clearpage  % TODO REVIEW/FINAL: This \clearpage needs to be removed from both review and camera-ready versions.


\section*{Acknowledgements}
This work is supported in part by the National Natural Science Foundation of China, under Grant (62302309, 62171248), Shenzhen Science and Technology Program (JCYJ20220818101014030, JCYJ20220818101012025), and the PCNL KEY project (PCL2023AS6-1).

\bibliographystyle{splncs04}
\bibliography{egbib}
\end{document}